\documentclass[onecolumn]{openhelix}
\usepackage{graphicx}
\usepackage{amsmath}
\usepackage{amssymb}
\usepackage{amsfonts}
\usepackage{tabularx}
\usepackage{float}
\usepackage{algorithm}
\usepackage{algorithmic}
\usepackage{array}
\usepackage{booktabs}
\usepackage{makecell}
\usepackage{wrapfig}
\usepackage{xspace}
\usepackage{nicefrac}
\usepackage{microtype}
\usepackage{multirow}
\usepackage{colortbl}
\usepackage{mathtools}
\usepackage{enumitem}

\usepackage[table]{xcolor}
\definecolor{lightgrey}{HTML}{dcdbdb}
\definecolor{lightblue}{HTML}{E8F0FE}
\definecolor{gray}{HTML}{9aa0a6}
\definecolor{lightpink}{HTML}{F48FB1}
\definecolor{lightred}{HTML}{FFCBC9}
\definecolor{lightcyan}{HTML}{80DEEA}

\definecolor{newyellow}{HTML}{FFD94D}
\definecolor{newgrey}{HTML}{7F7F7F}
\definecolor{newpink}{HTML}{FBCDF4}
\definecolor{realworldoft}{HTML}{029533}
\definecolor{realworldsf}{HTML}{8CC46A}
\definecolor{realworldour}{HTML}{FFC715}
\newcommand{\method}{DFM-VLA}

\title{DFM-VLA: Iterative Action Refinement for Robot Manipulation via Discrete Flow Matching}
\author[1*]{Jiayi Chen}
\author[1*]{Wenxuan Song}
\author[2]{Jiaxin Fang}
\author[2]{Ruiqing Yin}
\author[1]{Jingbo Wang}
\author[3,4]{Shuai Chen}
\author[1]{Jieyuan Pei}
\author[1]{Yikai Qin}
\author[1]{Feifan Chen}
\author[1]{Haodong Yan}
\author[1]{Zhide Zhong}
\author[1]{Wen Chen}
\author[5]{Yan Wang}
\author[2]{Yuxiang Gao}
\author[1\dagger]{Haoang Li}

\affiliation[1]{The Hong Kong University of Science and Technology (Guangzhou)}
\affiliation[2]{COCO Matrix}
\affiliation[3]{ShanghaiTech University}
\affiliation[4]{Shanghai Institute of Technical Physics, CAS}
\affiliation[5]{Tsinghua University, AIR}

\contribution[*]{Equal Contribution}
\contribution[\dagger]{Corresponding Author}

\abstract{
Vision--Language--Action (VLA) models that
encode actions using a discrete tokenization
scheme have been widely adopted for robotic
manipulation, but existing decoding paradigms
remain fundamentally limited.
Whether actions are decoded sequentially by
autoregressive VLAs or in parallel by discrete
diffusion VLAs, once a token is generated, it is
typically fixed and cannot be revised in
subsequent iterations. Consequently, early token
errors cannot be effectively corrected later.
We propose \method, a discrete flow matching
VLA that iteratively refines action tokens.
\method~models a token-level probability velocity
field that dynamically updates the full action
sequence across refinement iterations.
We investigate two approaches to constructing the
velocity field: an auxiliary velocity-head
formulation and an embedding-guided formulation.
To further improve prediction accuracy, we
introduce a metric-aligned action tokenizer
(MAAT) tailored to the coarse-to-fine nature of
DFM, together with a two-stage decoding strategy.
Extensive experiments on CALVIN, LIBERO,
LIBERO-Plus, and real-world manipulation tasks
demonstrate the effectiveness of our approach.
Our project is available at
\url{https://chris1220313648.github.io/DFM-VLA/}.
}

\begin{document}

% Keep the three-author OpenVLA-OFT citation compact in comparison tables.
\shortcites{kim2025fine}

\maketitle
\section{Introduction}

\begin{figure*}[t]  
    \centering
    \includegraphics[width=0.98\linewidth]{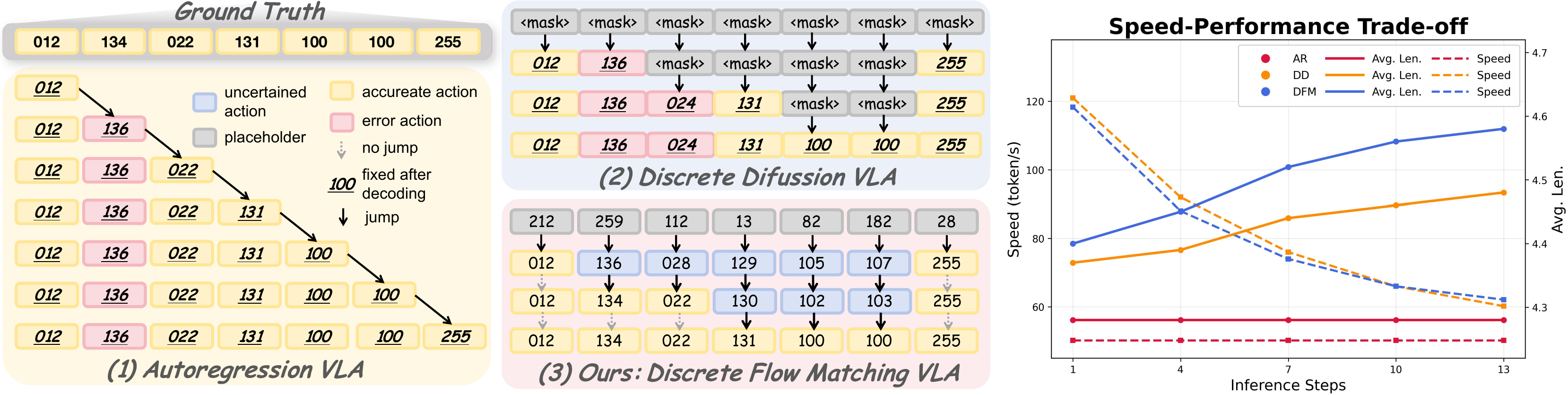} 
    \caption{Comparison of decoding paradigms. 1) Autoregressive (AR) models require as many steps as the action sequence length, while 2) Discrete Diffusion (DD) models enable faster generation through parallel token updates. However, both have the same limitation that once an erroneous token is produced, it cannot be corrected in later iterations.
    We refer to this phenomenon as \emph{irreversible commitment}.
    In contrast, 3) our Discrete Flow-matching (DFM) VLA  performs full-sequence action refinement at every iteration, allowing token-level correction and improving action quality for robotic manipulation.}
    \label{fig:decode_compare} 
    \vspace{-0.3cm}
\end{figure*}

Vision--Language--Action (VLA)~\citep{univla,Pi0,gr00t,cui2025openhelix,kim2024openvla,song2025reconvla,wang2025vlaadapter,zhong2026dualcot,yan2026svam} models have become a promising foundation for robotic manipulation, where policies map language instructions and visual observations to executable actions. 
A common and effective design is to discretize actions into tokens, which enables scalable training with Vision-Language Model (VLM) backbones and leverages their strong capabilities in vision and language
understanding.

Most existing discrete VLA systems fall into two families. Autoregressive (AR) methods~\citep{brohan2022rt,kim2024openvla,univla} decode tokens sequentially with a next-token prediction objective. 
However, their inherently left-to-right decoding order makes it difficult to revise erroneous tokens once they are emitted.
Discrete diffusion (DD) methods~\citep{song2025accelerating,liang2025discrete,wen2025llada} improve parallelism and often reduce latency, but they may still produce erroneous tokens in early iterations under confidence-guided decoding (see \Cref{fig:decode_compare}).
As a result, AR and DD paradigms struggle with a shared issue in robot control: early decoding errors cannot be effectively corrected and therefore propagate through the action chunk, degrading downstream robotic task performance.
% To alleviate this issue, \citet{liang2025discrete} introduce confidence-based remasking strategy. However, model confidence does not always reliably indicate prediction correctness~\citep{ni2026flexibility}, so this strategy may still fail to identify and revise erroneous actions.

Recent Discrete Flow Matching (DFM)~\citep{wang2025fudoki,luo2025next,havasi2025edit,nguyen2025oneflow,deng2025uniform} studies on Large-Language Models (LLMs) and VLMs have demonstrated competitive performance on text reasoning and editing, as well as image and video generation, by enabling iterative token refinement and flexible generation trajectories guided by a velocity field.
These advances inspire us to transfer the same refinement principle to discrete action generation for robotic control.
% Inspired by these advances, we aim to overcome the limitations of AR- and DD-based methods.

We propose \method, a VLA framework that unifies language, vision, and action in a discrete formulation and employs a discrete action velocity field to enable holistic refinement of action sequences.
As shown in \Cref{fig:decode_compare}, instead of treating action prediction as one-shot token generation, \method~models a token-level probability velocity field and performs iterative full-sequence refinement. 
This formulation allows the model to repeatedly revisit previously updated positions and selectively correct uncertain tokens as context becomes more informative.
For velocity-field construction, we explore two variants, namely an auxiliary velocity head formulation that predicts velocities from model hidden states, and an embedding-guided formulation that defines semantically structured probability paths in the
action-token embedding space and derives the
corresponding kinetic-optimal velocities.
To make discrete actions more amenable to velocity-based refinement, we further introduce a \emph{Metric-Aligned Action Tokenizer} (MAAT). Unlike conventional tokenizers, whose embedding spaces do not necessarily preserve numerical relationships between actions, MAAT discretizes continuous action scalars with a uniform codebook and aligns distances between token embeddings with the corresponding distances in the continuous action space. This metric-aware representation provides a meaningful geometry for probability-path construction, enabling stable coarse-to-fine action refinement under DFM.
For decoding, we adopt a two-stage strategy consisting of an iterative refinement stage for exploration and correction, followed by a validation stage for stable convergence.

We evaluate \method~on CALVIN, LIBERO, LIBERO-plus and real-world manipulation tasks. Across benchmarks, \method~consistently improves robot manipulation quality while maintaining strong inference efficiency. Ablation studies further indicate that the proposed design is robust to key tokenizer design, training data scale, and two-stage decoding allocation.

Our contributions are summarized as follows:
\begin{itemize}
    % \item We identify the \emph{irreversible commitment} problem in existing autoregressive and discrete diffusion VLA decoding, which limits token correction in robot manipulation.
    \item We propose \method, a discrete flow matching VLA framework that iteratively refines full action sequences through token-level velocity prediction, and systematically compare two strategies for constructing the velocity field.
    \item We introduce a metric-aligned action tokenizer together with a two-stage decoding strategy to further improve action prediction
accuracy.
    \item We demonstrate strong empirical performance on CALVIN, LIBERO, LIBERO-plus and real-world manipulation tasks, achieving consistent improvements.
\end{itemize}

\section{Related Works}
% \paragraph{Autoregressive VLA.}
% Autoregressive VLA models(see~\Cref{fig:decode_compare}) represent actions as discrete tokens, enabling LLM-style decoding and scalable training.
% Representative systems include RT-1~\citep{brohan2022rt}, OpenVLA~\citep{kim2024openvla}, and LLaVA-VLA~\citep{song2026rethinking}, which unify vision, language, and action prediction under the next-to-next objective.
% % Subsequent work improves perception or reasoning in discrete VLA pipelines, such as Robo-Flamingo~\citep{li2024vision}, GR-1~\citep{wu2023unleashing}, and 
% ReconVLA~\citep{song2025reconvla} improves perception by reconstructing the current frame to assist discrete action generation.
% Meanwhile, dynamic inference or efficiency tokenizer designs (e.g., Deer-VLA~\citep{yue2024deer} and FAST~\citep{pertsch2025fast}) target practical deployment constraints.
% Methods based on vector quantization (VQ)~\citep{wang2025vq,liu2025faster,dong2026actioncodec} provide a flexible alternative by learning discrete latent representations.
% UniVLA~\citep{univla} and WorldVLA~\citep{worldvla} discretize future frames to leverage dynamic visual information for action generation during both training and inference.
% % 
% These autoregressive models are effective but inherently sequential, which motivates our refinement alternative.
\paragraph{Discrete Diffusion VLA.}
Discrete diffusion VLA models extend dLLM-style~\citep{yu2025discrete} parallel denoising to action sequences, enabling full-attention, multi-token decoding with iterative refinement.
Representative methods differ in their corruption processes and efficiency strategies.
PD-VLA~\citep{song2025accelerating} follows a BART-style noising scheme by substituting tokens with vocabulary items and learning to reconstruct the sequence, while Discrete Diffusion VLA~\citep{liang2025discrete} and LLADA-VLA~\citep{wen2025llada} adopt BERT-style masking with a dedicated mask token~\citep{bert}.
To reduce inference cost, CEED-VLA~\citep{song2025ceed} applies consistency distillation to shrink the number of denoising steps with minimal performance loss.
On the data and modeling side, Dream-VLA~\citep{yedreamVLA} scales discrete diffusion pretraining on the OXE~\citep{o2024open} datasets following OpenVLA~\citep{kim2024openvla}.
Likewise, UD-VLA~\citep{chen2025unified}, dVLA~\citep{wen2025dvla}, and MM-ACT~\citep{liang2025mm} incorporate visual or textual chain-of-thought (CoT) and jointly diffuse future frames, reasoning traces, and action tokens for unified perception--reasoning--action generation.
However, many confidence-based decoding schemes commit high-confidence tokens early and provide limited opportunities for subsequent revision.
\paragraph{Discrete Flow Matching Models.}
Discrete flow matching~\citep{gat2024discrete} provides a principled framework for modeling probability paths over discrete tokens, with theoretical work~\citep{shaul2024flow} establishing kinetic-optimal discrete paths and velocity formulations that generalize mask-based mixtures.
Building on these foundations, several large-scale multimodal models adopt discrete flow matching for unified understanding and generation, including Fudoki~\citep{wang2025fudoki} and Next-Omni~\citep{luo2025next}, demonstrating strong any-to-any or omnimodal capabilities.
EditFlow~\citep{havasi2025edit} introduces explicit edit operations (insertion, replacement, deletion) with learned rates to enable flexible sequence refinement, while OneFlow~\citep{nguyen2025oneflow} extends this idea to concurrent mixed-modal and interleaved generation.
URSA~\citep{deng2025uniform} further illustrates how discrete metric structures can guide token transitions in generative modeling.
In autonomous driving, WAM-Flow~\citep{xu2025wam} extends discrete flow matching to Vision--Language--Navigation (VLN) by casting ego-trajectory planning as structured-token refinement.
We introduce discrete flow matching to the action modality, leveraging iterative refinement via a velocity field to progressively enhance action generation.

\section{Preliminary: Discrete Flow Matching}
\label{sec:preliminary}
We briefly review the key concepts and notation of discrete flow matching \citep{gat2024discrete,wang2025fudoki} that are used throughout the paper. 
DFM aims to transform a known source distribution $p(x)$ into a target data distribution $q(x)$ over a discrete space. 
We consider $x=(x^1, x^2, \ldots, x^D) \in \mathcal{S}=\mathcal{T}^D$, where $D$ is the number of discrete variables and $\mathcal{T}=[K]=\{1,2,\ldots,K\}$ is the finite alphabet of possible values.

\noindent
\textbf{Probability Paths}.
Given a \emph{source distribution} \( p(x) \) and a \emph{target distribution} \( q(x) \) on a finite state space \( \mathcal{S} \), DFM defines a family of time-indexed distributions \( \{p_t(x)\}_{t \in [0,1]} \) that smoothly interpolates between \( p \) and \( q \), referred to as \emph{probability paths}.
Each \( p_t(x) \) is constructed as
$p_t(x) \coloneqq \sum_{x_1 \in \mathcal{S}} p_t(x \mid x_1) q(x_1)$,
where the conditional distribution factorizes across dimensions, i.e., $p_t(x \mid x_1) \coloneqq \prod_{i=1}^D p_t(x^i \mid x_1^i)$.
Each term \( p_t(x^i \mid x_1^i) \) interpolates between the base distribution \( p(x^i) \) and a point mass \( \delta_{x_1^i}(x^i) \), i.e., $\delta_{x_1^i}(x^i)=1$ if $x^i=x_1^i$ and $0$ otherwise. 
% We impose the boundary conditions as $p_0(x^i \mid x_1^i) = p(x^i)$ and $\quad p_1(x^i \mid x_1^i) = \delta_{x_1^i}(x^i)$,
% ensuring the path starts from the factorized distribution \( p(x) = \prod_i p(x^i) \) and ends at the target \( q(x) \).
A common choice is the \emph{mixture path} used in~\citep{gat2024discrete,havasi2025edit}, defined by a time-dependent scheduler \( \kappa_t(x_1^i) \in [0, 1] \):
\begin{equation}
p_t(x^i \mid x_1^i) = (1 - \kappa_t(x_1^i)) p(x^i) + \kappa_t(x_1^i) \delta_{x_1^i}(x^i),
\label{eq:mask_probability_path}
\end{equation}
where \( \kappa_0(\cdot) = 0 \) and \( \kappa_1(\cdot) = 1 \). The term $p_t(x^i \mid x_1^i)$ is a \emph{conditional} forward probability path that describes how the state $x^i$ evolves given $x_1^i$. 

\noindent
\textbf{Probability velocities.}
To realize the prescribed probability path $p_{t}(x)$, we use a Continuous-Time Markov Chain (CTMC). Its dynamics are governed by a probability velocity $u_t$, also called the \emph{transition rate}, which describes how the current state $x_t$ moves toward the target state $x_1$ over time. Each token $i$ is updated independently according to
\begin{equation}\label{eq:sample}
x_{t+h}^i \sim \delta_{x_t^i}(\cdot) + h \, u_t^i(\cdot \mid x_t^i, x_1^i),
\end{equation}
where $u_t^i(\cdot \mid x_t^i, x_1^i)$ is the \emph{velocity field}, a conditional rate function that governs the flow of probability from $x_t^i$ to $x_1^i$.
Equation (\ref{eq:sample}) can be viewed as a small perturbation of the point mass $\delta_{x_t^i}$ scaled by the step size $h$, which models discrete state transitions in continuous time. The velocity field is central to DFM, which defines the probability-path dynamics and is the main quantity learned during training.

\section{Method}
\begin{wrapfigure}{r}{0.5\columnwidth}
    \centering
    \vspace{-0.5\baselineskip}
    \includegraphics[width=\linewidth]{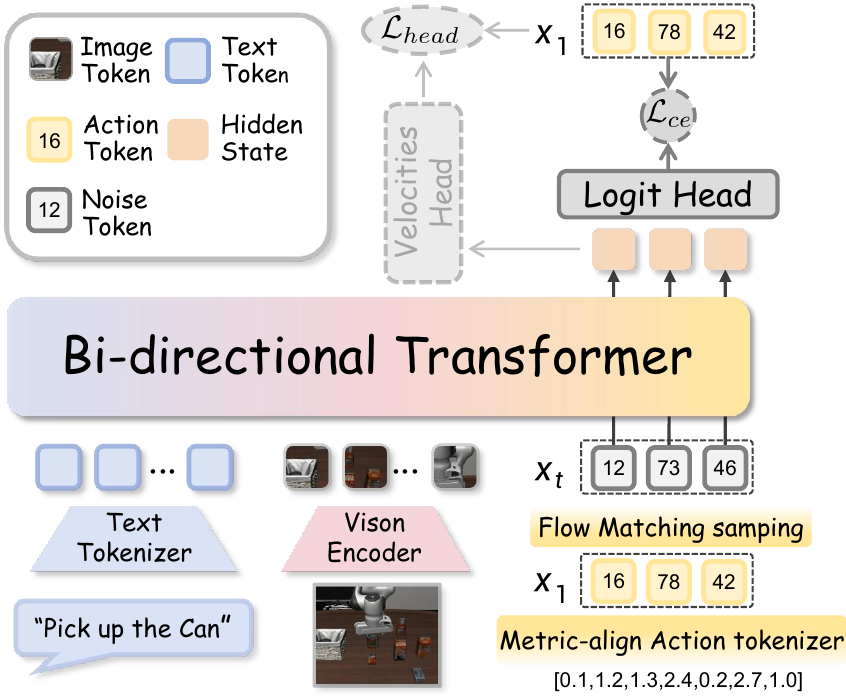}
    \caption{Overall architecture of~\method. Given language--vision context and noised action tokens $x_t$, the model predicts clean actions $x_1$ and learns the velocity field via loss function $\mathcal{L}_{\text{ce}}$ or $\mathcal{L}_{\text{head}}$.}
    \label{fig:model_arch}
    \vspace{-3\baselineskip}
\end{wrapfigure}
% This section is organized into four parts. We first introduce the model architecture. We then present two velocity-field constructions: 1) an auxiliary velocity head formulation and 2) an embedding-guided formulation. Finally, we describe the two-stage inference procedure.
This section first introduces the model architecture and the Metric-Aligned Action Tokenizer (MAAT), which provides a numerically structured action-token space. We then present two velocity-field constructions---an auxiliary-head formulation and an embedding-guided formulation---before describing the training and inference procedures.
\subsection{Architecture}
\label{subsec:architecture}
% 

% As shown in \Cref{fig:model_arch}, 
% DFM-VLA follows an architecture similar to UniVLA and adopts a unified discrete token formulation for language, vision, and action modalities.
As shown in \Cref{fig:model_arch}, 
DFM-VLA adopts a unified discrete token formulation for language, vision, and action modalities.
For the language modality, we tokenize the instructions with the Emu3~\citep{wang2024emu3} tokenizer.
For the visual modality, we discretize third-person and wrist-view observations with a VQ tokenizer~\citep{zheng2022movq} using a compression ratio of 4, representing each image as $25\times25=625$ tokens.
For the action modality, our base setting follows the FAST tokenizer~\citep{pertsch2025fast}, which compresses the action tokens with Byte Pair Encoding (BPE), yielding an action vocabulary size of 1024.
Furthermore, we introduce a metric-aligned action tokenizer that directly discretizes continuous action scalars using a uniform codebook while aligning distances between token embeddings with the corresponding distances in the continuous action space.
In this work, we denote the discretized instruction and observations as $l$, and apply noising and prediction only to the action modality.

% \noindent\textbf{Metric-Aligned Action Tokenizer (MAAT).}

\subsection{Metric-Aligned Action Tokenizer (MAAT)}
Unlike language tokens, action tokens represent ordered control values: nearby values should induce similar robot commands. A conventional embedding table, however, treats token identities as unrelated categories and provides no guarantee that this ordering is retained in the representation space. To encode this structure explicitly, MAAT quantizes each normalized robot action component on the shared grid
$\mathcal{V}_{\mathrm{tok}}=\{-1,-0.999,\ldots,1\}$,
which has a resolution of 0.001 and contains $N=2{,}001$ values. For a grid value $v_i$, we obtain its representation through a linear map $E:\mathbb{R}\to\mathbb{R}^d$ followed by L2 normalization, i.e., $z_i=E(v_i)/\lVert E(v_i)\rVert_2$.

We train this mapping to preserve the relative ordering of pairwise distances in the original action space. Given an anchor value $v_i$, we sample $v_j$ and $v_k$ such that $|v_i-v_j|<|v_i-v_k|$. Writing the corresponding embedding distances as $d_{ij}=\lVert z_i-z_j\rVert_2$ and $d_{ik}=\lVert z_i-z_k\rVert_2$, we optimize the triplet-margin objective
\begin{equation}
\label{eq:tok_loss}
\mathcal{L}_{\mathrm{tok}}
=\mathbb{E}_{(i,j,k)\sim\mathcal{T}}\big[\max\big(0,\; d_{ij}-d_{ik}+\gamma\big)\big],
\end{equation}
where $\mathcal{T}$ denotes the triplet sampling distribution and $\gamma$ is the margin. This objective constrains a numerically closer action value to remain closer in the embedding space which preserves relative distance order. The resulting monotonic neighborhood structure supplies the metric $d_i(\cdot,\cdot)$ used by the subsequent embedding-guided probability path.

\subsection{Modeling Velocities by Auxiliary Head}
\label{sec:head_vel}
% inspire by editflow，我们探索了构建速度场的另一种形式，editflow通过额外的速度头同时建模插入替换以及删除的三种速度场，我们只保留替换操作，因为相对文本而言，动作token的长度设计好的，相对于动作预测任务单独建模动作token的替换速率场就可以，
Drawing inspiration from EditFlow~\citep{havasi2025edit}, which employs auxiliary velocity heads to model three types of edit operations, namely \emph{insertion}, \emph{replacement}, and \emph{deletion}, we explore an alternative formulation for velocity-field construction. We retain only the \emph{replacement} operation, because the action token sequence length is predefined by our action chunk design and the three operations are fundamentally equivalent under appropriate reformulation~\citep{nguyen2025oneflow,havasi2025edit}.
% Consequently, for action prediction, modeling only a replacement velocity field is sufficient.
Concretely, given noisy action tokens $x_t$ and context $l$, the backbone first produces hidden states, and the auxiliary velocity head then maps these hidden states to velocities:
\begin{equation}
\label{eq:head-velocity}
h_t = f_{\theta}(x_t, l), \qquad
u_t^{\theta}(\cdot \mid x_t) = u_t^{\text{head}}(h_t),
\end{equation}
where $f_{\theta}$ denotes the backbone network and $u_t^{\text{head}}$ denotes the velocity prediction head.

\textbf{Loss Function.}
We train the auxiliary velocity head using the following velocity-matching objective:
\begin{equation}
\label{eq:head_loss}
\begin{aligned}
\mathcal{L}_{\text{head}}
\!&=
\mathbb{E}_{t \sim \mathcal{U}[0,1], {x}_1, {x}_t}
\bigg[
\sum_{x\neq x_t} u_t^\theta(x\mid x_t) - \\
&\qquad 
\sum_{i=1}^{D}\mathbf{1}_{[x^i_t\neq x^i_1]}\,
% \frac{\dot{\kappa}_t}{1-\kappa_t}\,
\log u_t^\theta\!\left(x^i\mid x^i_t\right)
p_{1 \mid t}\left( {x}^i_1 \mid {x}^i_t, {l}\right)
\bigg].
\end{aligned}
\end{equation}
% \begin{equation}
% \mathcal{L_\text{head}}
% =
% \mathbb{E}_{t \sim \mathcal{U}[0,1],\, }
% \left[
% \sum u_t^\theta(x\mid x_t)
% -
% \sum_{i=1}^{D}\mathbf{1}_{[x^i_t\neq x^i_1]}\,
% % \frac{\dot{\kappa}_t}{1-\kappa_t}\,
% \log u_t^\theta\!\left(x^i\mid x^i_t\right)p_{1 \mid t}\left( {x}^i_1 \mid {x}^i_t, {l}\right)
% \right].
% \end{equation}
\noindent Here, \(\mathbf{1}_{[x_t^i \neq x_1^i]}\) is an indicator whether the current token differs from the target, encouraging higher flow velocities for tokens that still require refinement.

\subsection{Embedding-Guided Velocity Modeling}
\label{sec:Embed_vel}

Building on recent advances in discrete flow matching~\citep{gat2024discrete,wang2025fudoki,deng2025uniform}, we parameterize the probability path in a metric-induced form. 
Specifically, let \( d : \mathcal{T} \times \mathcal{T} \to \mathbb{R}_{\geq 0} \) be a distance such that \( d(x^i, x_1^i) = 0 \) if and only if \( x^i = x_1^i \), where \( d(\cdot, \cdot) \) is measured in the action token embedding space. For metric-aligned action tokens, this distance is computed from the action-token embeddings trained by \(\mathcal{L}_{\mathrm{tok}}\) in Eq.~\ref{eq:tok_loss}, and the resulting coordinate-wise distance is used as \(d_i(\cdot,\cdot)\). We define the conditional path as
\begin{equation}
p_t(x^i \mid x_1^i) = \mathrm{softmax}\big( -\beta_t \cdot d(x^i, x_1^i) \big),
\label{eq:diffusion-softmax}
\end{equation}
where $\beta_t:[0,1] \to \mathbb{R}_{\ge 0}$ is a monotonic schedule with boundary conditions $\beta_0 = 0$ and $\beta_1 = \infty$. We instantiate it as
\begin{align}\label{eq:beta_t}
\beta_t = c \left(\dfrac{t}{1-t}\right)^\alpha, \qquad t \in [0,1),
\end{align}
where $c>0$ and $\alpha>0$ control how fast probability mass concentrates toward the target token over time. 
% Compared with the mask-based path in Eq.~\ref{eq:mask_probability_path}, 
This formulation preserves semantic neighborhood structure such that tokens closer to $x_1^i$ receive larger probability as $t \to 1$.

Given this prescribed path, we adopt the kinetic-optimal velocity obtained by minimizing transport energy under the flow constraints~\citep{wang2025fudoki,deng2025uniform,luo2025next}:
\begin{equation}
\begin{aligned}
u_t^i(x^i, z \mid x_1) = p_t(x^i \mid x_1^i) \, \dot{\beta}_t \, [ d(z^i, x_1^i) - d(x^i, x_1^i) ]_+
\end{aligned} 
\label{eq:ko-velocity}
\end{equation}
where $[\cdot]_+ = \max\{\cdot, 0\}$,  $z^i$ is a token in the vocabulary $\mathcal{T}$, and $\dot{\beta}_t$ is the derivative of $\beta_t$ w.r.t. $t$. This velocity moves probability mass from $z^i$ to $x^i$ only when $x^i$ is closer to $x_1^i$ than $z^i$, yielding a monotonic refinement process toward the clean target token.

\noindent
\textbf{Loss Function.}
Given corrupted action tokens $x_t$, the model predicts the target action sequence $x_1$ by outputting per-position categorical logits. We optimize the expected cross-entropy:
\begin{equation}
\label{eq:training_loss}
\mathcal{L}_{\text{ce}} = \mathbb{E}_{t \sim \mathcal{U}[0,1],\, {x}_1, {x}_t}
\left[ -\log p_{1 \mid t}\left( {x}_1 \mid {x}_t, {l}\right) \right].
\end{equation}
Here, $p_{1|t}^{\theta}(\cdot \mid x_t, l)$ denotes the predicted categorical distribution of the model at each action token position.

% As shown in~\Cref{fig:velocity_field_compare}, the embedding-guided objective (\(\mathcal{L}_{\text{ce}}\)) converges faster and achieves better final performance than the head-based objective (\(\mathcal{L}_{\text{head}}\)). Unless otherwise specified, all experiments use the embedding-guided setting.

\subsection{Training}
\label{subsec:training}
We adopt a two-stage training pipeline. In the first stage, we jointly train the metric-aligned action tokenizer and the VLM for two epochs on a mixture of LIBERO and CALVIN datasets, using $\mathcal{L}_{\mathrm{tok}}$ in~\Cref{eq:tok_loss} together with $\mathcal{L}_{\mathrm{ce}}$. In the second stage, we freeze MAAT and fine-tune only the VLM for four epochs on each downstream benchmark, using $\mathcal{L}_{\mathrm{ce}}$ or $\mathcal{L}_{\text{head}}$.

\begin{algorithm}[t]
\footnotesize
\caption{Two-Stage Decoding of \method}
\label{alg:two_stage_sampling}
\begin{algorithmic}[1]
\STATE \textbf{Input:} predictor $p_\theta$, context $l$, steps $T_{\mathrm{fine}},T_{\mathrm{val}}$, action vocabulary $\mathcal{V}$
\STATE Sample $x_0 \sim \mathrm{Uniform}(\mathcal{V})$; set $T \gets T_{\mathrm{fine}} + T_{\mathrm{val}}$
\FOR{$k=1$ to $T$}
    \STATE $t \gets (k-1)/T$, $h \gets 1/T$
    \STATE $\hat{x}_1 \sim p_\theta(\cdot \mid x_t, l)$
    \IF{$k \le T_{\mathrm{fine}}$}
        \STATE Compute velocity $u_t$ from $\hat{x}_1$ (Eq.~\ref{eq:ko-velocity} or Eq.~\ref{eq:head-velocity})
        \STATE Update $x_{t+h}$ by a  CTMC Euler step
    \ELSE
        \STATE Update $x_{t+h} \gets \arg\max p_\theta(\cdot \mid x_t, l)$
    \ENDIF
\ENDFOR
\STATE \textbf{Output:}  action sequence $x_1$
\end{algorithmic}
\end{algorithm}
\subsection{Inference}
\label{subsec:inference}

\begin{figure}[t]  
    \centering
    \includegraphics[width=0.9\linewidth]{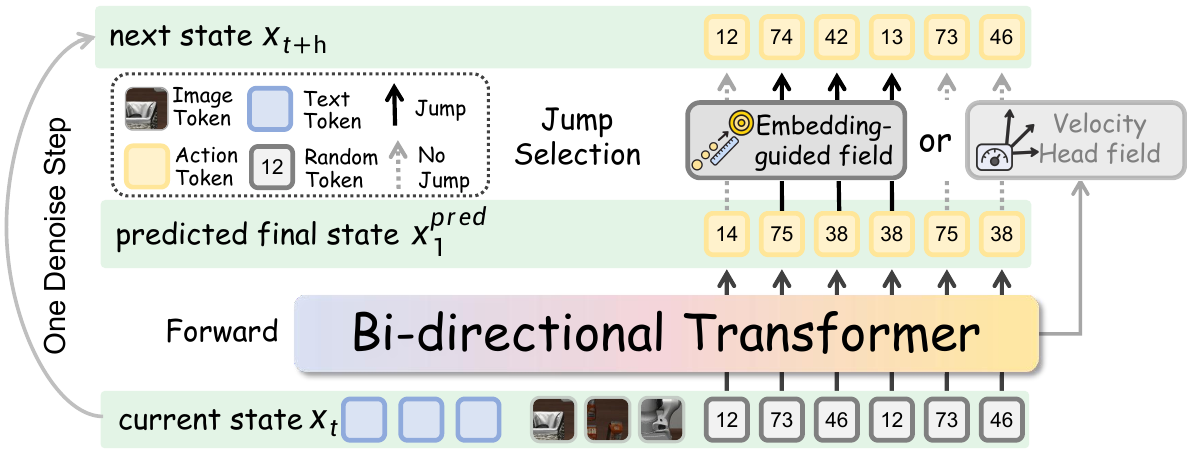} 
    \caption{Visualization of a single decoding step in the iterative refinement stage. After predicting final state $x^\text{pred}_1$, the model does not directly output final action tokens. Instead, it constructs a velocity field to compute transition rates and selectively updates tokens to next state $x_{t+h}$ at each step.}
    \label{fig:inference} 
     \vspace{-0.3cm}
\end{figure}

We perform inference in two stages: an iterative refinement stage followed by a deterministic validation stage, with $T_{\mathrm{fine}}$ and $T_{\mathrm{val}}$ decoding steps, respectively.

In the iterative refinement stage, we employ an Euler discretization of the continuous-time Markov chain process \( (x_t)_{0 \leq t \leq 1} \), following the approach in~\citep{deng2025uniform}. As illustrated in \Cref{fig:inference}, for each coordinate \(i\) and each step from \(t\) to \(t+h\), we perform:
\begin{itemize}
  \item Sample \( x_1^i \sim p_{1|t}^i(\cdot \mid x_t, l) \) from the model;
  \item Compute the total outgoing rate \( \lambda^i = \sum_{x^i \neq x_t^i} u_t^i(x^i, x_t^i \mid x_1^i) \) using Eq.~\ref{eq:ko-velocity} or Eq.~\ref{eq:head-velocity};
  \item Draw \( Z^i_{\text{change}} \sim U[0,1] \);
  \item Update \( x_{t+h}^i \): if \( Z^i_{\text{change}} \leq 1 - e^{-h\lambda^i} \), sample from \( \frac{u_t^i(\cdot, x_t^i \mid x_1^i)}{\lambda^i}(1-\delta_{x_t^i}(\cdot)) \); otherwise keep \( x_{t+h}^i=x_t^i \).
\end{itemize}
Here, \(\lambda^i\) is the total transition intensity out of the current token \(x_t^i\), so the jump probability \(1-e^{-h\lambda^i}\) increases with \(\lambda^i\). When a jump occurs, normalized rates favor states with larger velocity flow, which typically move the token closer to the predicted clean state \(x_1^i\). Repeating this process over time enables iterative correction across the full sequence.
Compared with mask-based discrete diffusion methods~\citep{yedreamVLA}, this decoding scheme allows previously updated tokens to be revised again in later iterations, rather than being permanently fixed after being predicted.
% \looseness=-1

\noindent
During the \textbf{validation} stage, we adopt a greedy decoding strategy to improve stability in the final refinement steps.
In the final $T_{\mathrm{val}}$ decoding steps, we disable stochastic jumps and switch to greedy decoding,
% \begin{equation}
%     x_{t+h} = \arg\max \mathrm{softmax}(p_{1|t}(\cdot | x_t)),
% \end{equation}
\begin{equation}
x_{t+h}^{i}
=
\underset{a \in \mathcal{V}}{\arg\max}\;
p_{1|t}^{\theta,i}\!\left(a \mid x_t,l\right).
\end{equation}
% where $\hat{y}_t$ is the model output logits. 
This hybrid design preserves exploratory refinement in earlier iterations while enforcing deterministic convergence near the end, leading to more stable final-stage action predictions and improved reproducibility. 
The concise two-stage decoding procedure is summarized in \Cref{alg:two_stage_sampling}.

\noindent
\textbf{Adaptive KV Caching.}
Following the dynamic caching strategy of FAST-dLLM~\citep{wu2025fast} on discrete diffusion decoding, we exploit the fact that many tokens exhibit only minor KV-state changes across iterative denoising steps of DFM. 
We keep the instruction and observation KV caches largely fixed throughout inference, while adaptively updating the action-side cache based on the cosine similarity between current and cached value features.
Combined with the parallel refinement of \method, this dynamic KV reuse yields a 2.4$\times$ latency speedup over autoregressive decoding while preserving task performance (see~\Cref{tab:efficiency_main}).

\begin{table}[t]
    \footnotesize
    \setlength{\tabcolsep}{5pt}
    \centering
    \caption{\textbf{Comprehensive Evaluation of Long-Horizon Robotic Manipulation on the CALVIN Benchmark.} \textit{w/o Embed} denotes constructing the velocity field with the auxiliary head.}
    \resizebox{\columnwidth}{!}{%
    \begin{tabular}{l c c c c c c}
        \toprule
        \multirow{2}{*}{\textbf{Method}} & \multicolumn{5}{c}{\textbf{Tasks Completed in a Row}} & \multirow{2}{*}{\textbf{Avg. Len. $\uparrow$}} \\
        \cmidrule(lr){2-6}
        & 1 & 2 & 3 & 4 & 5 & \\
        \midrule
        % MCIL~\citep{lynch2020language} & 0.373 & 0.027 & 0.002 & 0.000 & 0.000 & 0.40\\
        RT-1~\cite{brohan2022rt} & 0.844 & 0.617 & 0.438 & 0.323 & 0.227 & 2.45 \\
        Robo-Flamingo~\citep{li2024vision} & 0.964 & 0.896 & 0.824 & 0.740 & 0.660 & 4.09 \\
        Deer~\citep{yue2024deer} & \textbf{0.982} & 0.902 & 0.821 & 0.759 & 0.670 & 4.13 \\
        GR-1~\citep{wu2023unleashing} & 0.949 & 0.896 & 0.844 & 0.789 & 0.731 & 4.21 \\
        ReconVLA~\citep{song2025reconvla} & \underline{0.980} & 0.900 & 0.845 & 0.785 & 0.705 & 4.23 \\
        UniVLA$^{*}$~\citep{univla} & 0.948 & 0.906 & 0.862 & 0.834 & 0.690 & 4.24 \\
        MODE~\citep{reussefficient} & 0.971 & 0.925 & 0.879 & 0.835 & 0.779 & 4.39 \\
        UP-VLA~\citep{zhang2025upvla} & 0.962 & 0.921 & 0.879 & 0.842 & \textbf{0.812} & 4.42 \\
        % MDT~\citep{reuss2024multimodal} & 0.986 & 0.958 & 0.916 & 0.862 & 0.801 & 4.52 \\
        % RoboVLMs~\citep{li2024towards} & 0.967 & 0.930 & 0.899 & 0.865 & 0.826 & 4.49 \\
        \rowcolor[gray]{0.9} \textbf{\method~w/o Embed} & 0.968 & 0.928 & 0.880 & \textbf{0.864} & 0.776 & 4.42\\
        \rowcolor[gray]{0.9} \textbf{\method~w/o MAAT} & 0.976 & \underline{0.944} & \underline{0.892} & 0.844 & 0.780 & \underline{4.44}\\
        \rowcolor[gray]{0.9} \textbf{\method} & 0.989 & \textbf{0.967} & \textbf{0.927} & \underline{0.872} & \underline{0.822} & \textbf{4.58}\\
        \bottomrule
    \end{tabular}
    }
    \label{tab:calvin_results}
    \vspace{-0.3cm}
\end{table}

\begin{table*}[h]
    \centering
    \footnotesize
    \setlength{\tabcolsep}{2.5pt}
    \caption{Evaluation and comparison on the LIBERO and LIBERO-Plus benchmarks. All values are success rates (\%).}
    \label{tab:libero}
    \label{tab:libero_plus}
    \resizebox{0.99\textwidth}{!}{%
    \begin{tabular}{lcccc|c||ccccccc|c}
        \toprule
        \multirow{2}{*}{\textbf{Method}}
        & \multicolumn{5}{c}{\textbf{LIBERO}}
        & \multicolumn{8}{c}{\textbf{LIBERO-Plus}} \\
        \cmidrule(lr){2-6}\cmidrule(lr){7-14}
        & Spatial & Object & Goal & Long & Average
        & Camera & Robot & Language & Light & Background & Noise & Layout
        & Total \\
        \midrule
        OpenVLA~\citep{kim2024openvla}
        & 84.7 & 88.4 & 79.2 & 53.7 & 76.5
        & 0.8 & 3.5 & 23.0 & 8.1 & 34.8 & 15.2 & 28.5 & 15.6 \\
        WorldVLA~\citep{worldvla}
        & 87.6 & 96.2 & 83.4 & 60.0 & 81.8
        & 0.1 & 27.9 & 41.6 & 43.7 & 17.1 & 10.9 & 38.0 & 25.0 \\
        $\pi_0$-Fast~\citep{pertsch2025fast}
        & 96.4 & 96.8 & 88.6 & 60.2 & 85.5
        & 65.1 & 21.6 & 61.0 & 73.2 & 73.2 & 74.4 & 68.8 & 61.6 \\
        FlowVLA~\citep{zhong2025flowvla}
        & 93.2 & 95.0 & 91.6 & 72.6 & 88.1
        & 51.0 & 25.4 & 70.2 & 77.6 & 79.4 & 58.3 & 47.0 & 56.2 \\
        DreamVLA~\citep{dreamvla25}
        & 97.5 & 94.0 & 89.5 & 89.5 & 92.6
        & 26.2 & 17.6 & 67.0 & 77.5 & 91.6 & 53.6 & 43.5 & 48.9 \\
        OpenVLA-OFT~\citep{kim2025fine}
        & 96.2 & 98.3 & 96.2 & 90.7 & 95.3
        & 56.4 & 31.9 & 79.5 & 88.7 & \underline{93.3} & 75.8 & \underline{74.2} & 69.6 \\
        UniVLA~\citep{univla}
        & 95.4 & \underline{98.8} & 93.6 & 94.0 & 95.5
        & 1.8 & 46.2 & 69.6 & 69.0 & 81.0 & 21.2 & 31.9 & 42.9 \\
        % F1~\citep{f1_vla_2025}
        % & 98.2 & 97.8 & 95.4 & 91.3 & 95.7
        % & -- & -- & -- & -- & -- & -- & -- & -- \\
        $\pi_0$~\citep{Pi0}
        & 96.8 & \underline{98.8} & 95.8 & 85.2 & 94.2
        & 13.8 & 6.0 & 58.8 & 85.0 & 81.4 & 79.0 & 68.9 & 53.6 \\
        DD-VLA~\citep{liang2025discrete}
        & 97.2 & 98.6 & 97.4 & 92.0 & 96.3
        & 69.2 & 40.5 & 77.2 & 85.4 & 86.9 & 77.5 & 65.1 & 70.2 \\
        Fast-dVLA~\citep{song2026fastdvla}
        & 97.0 & 97.6 & \underline{98.8} & 92.8 & 96.6
        & 69.7 & 41.5 & 78.3 & 86.1 & 87.5 & 78.1 & 65.1 & 70.8 \\
        $\pi_{0.5}$~\citep{intelligence2025pi_}
        & \textbf{98.8} & 98.2 & 98.0 & 92.4 & 96.8
        & 70.3 & 41.7 & \underline{81.1} & \textbf{97.3} & \textbf{94.6} & 71.8 & \textbf{84.9} & \underline{75.7} \\
        RIPT-VLA~\citep{tan2025interactive}
        & \underline{98.6} & 98.6 & \textbf{99.0} & 93.8 & \underline{97.2}
        & 55.2 & 31.2 & 77.6 & 88.4 & 91.6 & 73.5 & \underline{74.2} & 68.4 \\
        \rowcolor[gray]{0.9} \textbf{\method~w/o Embed}
        & 94.2 & 96.4 & 92.8 & 90.4 & 93.5
        & 66.4 & 44.2 & 76.9 & 82.2 & 80.0 & 77.2 & 64.1 & 69.2 \\
        \rowcolor[gray]{0.9} \textbf{\method~w/o MAAT}
        & 97.8 & \underline{98.8} & 95.8 & \underline{94.2} & 96.7
        & \underline{70.6} & \underline{48.4} & \underline{81.1} & 86.4
        & 84.2 & \underline{81.4} & 68.3 & 73.4 \\
        \rowcolor[gray]{0.9} \textbf{\method}
        & 98.4 & \textbf{99.2} & 98.0 & \textbf{96.2} & \textbf{98.0}
        & \textbf{75.0} & \textbf{52.8} & \textbf{85.5} & \underline{90.8}
        & 88.6 & \textbf{85.8} & 72.7 & \textbf{77.8} \\
        \bottomrule
    \end{tabular}%
    }
    \vspace{-0.4cm}
\end{table*}

\section{Experiments}
We conduct comprehensive experiments to evaluate the effectiveness of \method~on both simulation benchmarks and real-world robotic manipulation tasks.
Our experiments are designed to answer the following research questions:

\noindent
\textbf{(RQ1)} How does \method~compare with recent state-of-the-art VLA methods on CALVIN and LIBERO benchmarks? (\Cref{sec:sota})

\noindent
\textbf{(RQ2)} What empirical insights can guide key design choices for \method, including two-stage decoding allocation and velocity-field construction? (\Cref{sec:ablation})

\noindent
\textbf{(RQ3)} What additional insights can we gain from in-depth analysis of decoding behavior, execution quality, and efficiency trade-offs? (\Cref{sec:analysis})

\noindent
\textbf{(RQ4)} Can \method~generalize effectively to real-world manipulation tasks? (\Cref{sec:real})

\subsection{Setup}
% \textbf{Implementation Details.}
% 我们采用与univla相同的backbone Emu3~

% For action modeling, we follow FAST and apply the Discrete Cosine Transform (DCT) to map continuous action trajectories into discrete action tokens.

We initialize model from checkpoints pretrained on robotic video data~\citep{univla}.
Unless otherwise specified, we use a learning rate of $1\times10^{-4}$ and a batch size of 8.
All training and inference are conducted on 8 NVIDIA H100 GPUs.
For simulation benchmarks (CALVIN and LIBERO), we train for 20k--32k steps depending on the setting, while each real-world task is trained for 5k steps.
Unless otherwise noted, following~\citep{luo2025next}, we set the noise-schedule parameters to $c=3$ and $\alpha=1$. Further analysis of these parameters is provided in the supplementary files.
% We then conduct ablation studies on $c$ and $\alpha$ to quantify their effects on iterative refinement.

\begin{figure}[t]
    \centering
    \includegraphics[width=0.8\linewidth]{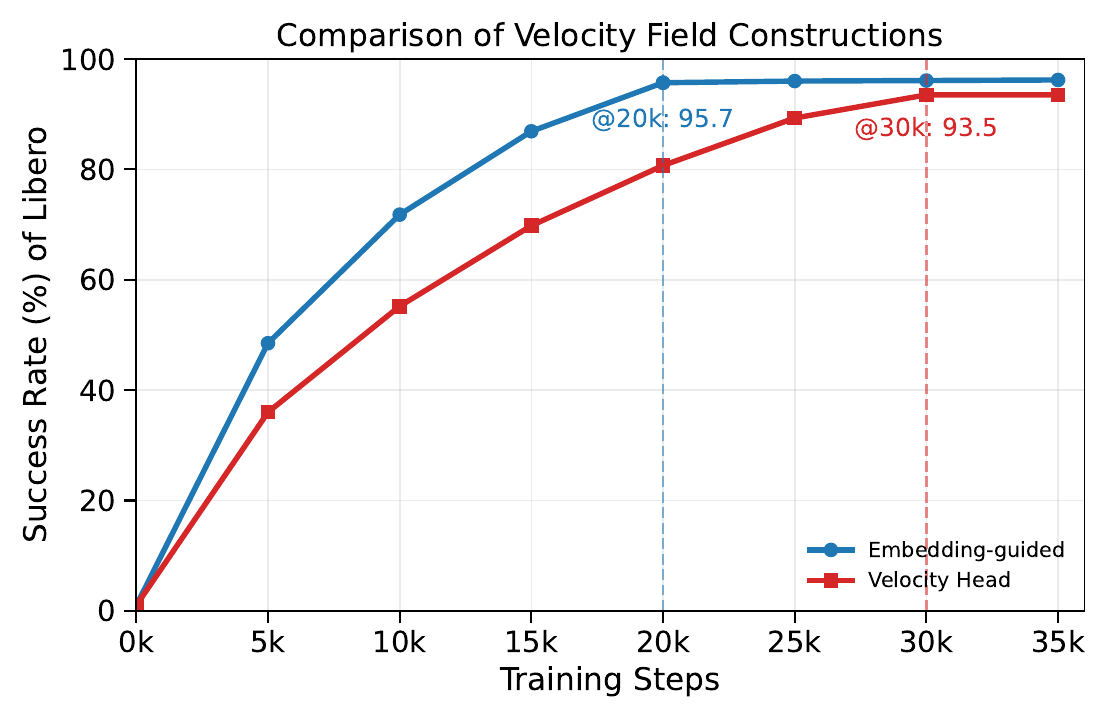}
    \caption{Comparison of velocity field constructions across training steps on LIBERO. 
    % The Embedding-guided variant consistently outperforms the Velocity Head variant in both convergence speed and final performance, reaching a strong success rate of 95.7\% on LIBERO with only 20k training steps.
    }
    \label{fig:velocity_field_compare}
    \vspace{-0.2cm}
\end{figure}

\subsection{Benchmarks}
\label{sec:benchmark}
\textbf{CALVIN.} CALVIN~\citep{mees2022calvin} evaluates long-horizon, language-conditioned manipulation across four environments (A--D), 34 skills, and 1,000 language annotations. We follow the standard \textbf{ABCD$\rightarrow$D} setup, in which each rollout comprises five consecutive language-conditioned sub-tasks whose later successes depend on earlier execution. We evaluate 1,000 rollouts per model and report per-step completion rates and the average number of consecutively completed sub-tasks (Avg. Len., maximum 5). 
\textbf{LIBERO.} LIBERO~\citep{liu2023libero} evaluates cross-task generalization through four suites: Spatial, Object, Goal, and Long, which respectively emphasize spatial relations, object generalization, goal-conditioned manipulation, and long-horizon composition. Each suite contains 10 tasks, and we conduct 50 rollouts per task and report the per-suite success rates and their average. 
\textbf{LIBERO-Plus.} LIBERO-Plus~\citep{fei2026liberoplus} extends LIBERO into 10,030 controlled perturbation tasks spanning seven robustness dimensions: camera viewpoints, robot initial states, language instructions, lighting conditions, background textures, sensor noise, and object layouts. Following official protocol, each perturbed task is evaluated with one rollout. We report the success rate for each perturbation dimension and the overall success rate across all tasks.

\subsection{Comparison with SOTA (RQ1)}
\label{sec:sota}

% We report the main results on CALVIN, LIBERO, and LIBERO-Plus in \Cref{tab:calvin_results,tab:libero}. 
As shown in  \Cref{tab:calvin_results}, \method~achieves the best average length of 4.58, outperforming the strongest unified diffusion VLA, UP-VLA, by 0.16. It also obtains the highest 2-step and 3-step completion rates of 0.967 and 0.927, respectively. Compared with the unified autoregressive UniVLA$^{*}$, \method~improves 3-step completion from 0.862 to 0.927 and 5-step completion from 0.690 to 0.822. These gains indicate that iterative action refinement reduces error accumulation and improves long-horizon consistency. The w/o Embed and w/o MAAT variants score 4.42 and 4.44, respectively, versus 4.58 for the full model, confirming their complementary benefits.

% We also observe that \method+Head attains the best 4-task completion rate (0.864), suggesting that explicit velocity-head modeling brings additional robustness in mid-horizon planning.

As shown in  \Cref{tab:libero}, \method~achieves the best average success rate of 98.0\%.
It exceeds the strongest continuous flow-matching baseline, $\pi_{0.5}$, by 1.2 percentage points and the strongest discrete-diffusion baseline, Fast-dVLA, by 1.4 points.
Compared with the  autoregressive baseline, UniVLA, the gain is 2.5 points. 
At the suite level, \method~obtains the best Object and Long success rates, reaching 99.2\% and 96.2\%, while remaining competitive on Spatial (98.4\%) and Goal (98.0\%).
This balanced performance across short-horizon and compositional suites leads to the strongest overall result.

On the more challenging LIBERO-Plus benchmark, \method~achieves the best total success rate of 77.8\%, outperforming $\pi_{0.5}$ by 2.1 percentage points.
% It ranks first under Camera, Robot, Language, and Noise perturbations, with success rates of 75.0\%, 52.8\%, 85.5\%, and 85.8\%, respectively.
Notably, \method~ranks first under language perturbations, achieving a success rate of 85.5\%. We attribute this robustness to the unified discrete formulation, which better preserves the semantic priors inherited from the pretrained VLM.
The full model outperforms its w/o Embed and w/o MAAT variants by 8.6 and 4.4 percentage points, respectively, showing that both components improve robustness under distribution shifts. 
Together, the LIBERO and LIBERO-Plus results show that discrete flow matching provides both strong in-distribution task performance and improved robustness to diverse visual, linguistic, and embodiment perturbations.

\subsection{Ablation Studies (RQ2)}
\label{sec:ablation}
We conduct ablation studies to isolate the effects of key design choices, including two-stage decoding step allocation and velocity-field construction. 
% Detailed quantitative results are summarized in \Cref{tab:ablation_two_stage,fig:velocity_field_compare}.

\begin{table}[t]
    \centering
    \begin{minipage}[t]{0.45\linewidth}
        \centering
        \scriptsize
        \setlength{\tabcolsep}{1.5pt}
        \captionof{table}{Two-stage decoding allocation ($T_{\mathrm{fine}}+T_{\mathrm{val}}=16$).}
        \label{tab:ablation_two_stage}
        \begin{tabular}{cccc}
            \toprule
            $T_{\mathrm{fine}}$ & $T_{\mathrm{val}}$ & CAL. $\uparrow$ & LIB. $\uparrow$ \\
            \midrule
            16 & 0  & 4.47 & 96.8 \\
            15 & 1  & 4.49 & 97.2 \\
            \rowcolor[gray]{0.9}
            14 & 2  & \textbf{4.58} & \textbf{98.0} \\
            12 & 4  & 4.53 & 96.4 \\
            % 8  & 8  & 4.35 & 95.1 \\
            \bottomrule
        \end{tabular}
    \end{minipage}
    \hfill
    \begin{minipage}[t]{0.48\linewidth}
        \centering
        \scriptsize
        \setlength{\tabcolsep}{1.5pt}
        \captionof{table}{Comparison of decoding strategies on CALVIN. }
        \label{tab:efficiency_main}
        \begin{tabular}{lcc}
            \toprule
            Method & Avg. $\uparrow$ & Speed (tokens/s)$\uparrow$ \\
            \midrule
            AR & 4.28 & 50.2 \\
            DD & 4.42 & 62.1 \\
            % DD + Adap. Cache & 4.27 & 118.3 \\
            DFM & \textbf{4.60} & 60.2 \\
            \rowcolor[gray]{0.9}
            DFM+Cache & 4.58 & \textbf{121.0} \\
            \bottomrule
        \end{tabular}
    \end{minipage}
    \vspace{-0.4cm}
\end{table}
\paragraph{Effect of Two-Stage Decoding.}
We further ablate how decoding steps are allocated between the iterative refinement stage ($T_{\mathrm{fine}}$) and the deterministic validation stage ($T_{\mathrm{val}}$). We keep the total number of steps fixed at 16 and report Avg. Len. of CALVIN ABCD$\rightarrow$D  and average success rate of LIBERO in \Cref{tab:ablation_two_stage}. As shown in the table, using only iterative refinement  ($T_{\mathrm{val}}=0$) yields weaker performance. Introducing a short validation stage improves both benchmarks, and $T_{\mathrm{fine}}=14, T_{\mathrm{val}}=2$ achieves the best overall trade-off. In contrast, allocating too many steps to the validation stage slightly hurts action quality, suggesting that excessive early greedy release reduces refinement flexibility. Therefore, we use $T_{\mathrm{fine}}=14$ and $T_{\mathrm{val}}=2$ in all experiments.

\paragraph{Comparison of Velocity Field Constructions.}
To further analyze the effect of velocity-field design, we compare the embedding-guided formulation with the head-based formulation across training steps. As shown in~\Cref{fig:velocity_field_compare}, the embedding-guided variant converges faster in the early stages of training and consistently achieves better task performance. This trend indicates that embedding guidance provides more informative and smoother optimization signals, leading to both improved data efficiency and a stronger policy.

\subsection{In-Depth Analysis (RQ3)}
\label{sec:analysis}

% \begin{wraptable}{r}{0.5\textwidth}
%     \vspace{-10mm}
%     \centering
%     \footnotesize
%     \setlength{\tabcolsep}{2pt}
%     \caption{Inference efficiency comparison across AR, DD and DFM on CALVIN ABCD$\rightarrow$D. DD and DFM use the same decoding steps (16).}
%     \label{tab:efficiency_main}
%     \begin{tabular}{lcc}
%         \toprule
%         Method & Avg. Len. $\uparrow$ & Speed $\uparrow$ \\
%         \midrule
%         AR & 4.18 & 50.2  \\
%         DD & 4.32 & 62.1  \\
%         DD + Adap. Cache & 4.27 & 118.3  \\
%         DFM & 4.58 & 60.2  \\
%         DFM + Adap. Cache & 4.40 & 121.0  \\
%         \bottomrule
%     \end{tabular}
%     \vspace{-6mm}
% \end{wraptable}
\paragraph{Effectiveness of Decoding Method.}

We keep the architecture unchanged and compare AR, DD, and DFM methods under the same evaluation setting. We further equip DFM with Adaptive Cache to assess its effect on inference efficiency.
As shown in \Cref{tab:efficiency_main}, DFM achieves the best action quality without caching, reaching an average length of 4.60. Adding Adaptive Cache increases its decoding speed from 60.2 to 121.0 while retaining an average length of 4.58, making \textit{DFM+Cache} the fastest variant. These results demonstrate that DFM offers a favorable quality--efficiency trade-off and can be substantially accelerated with minimal performance degradation.

\paragraph{Effect of training data size.}
% \vspace{-0.3cm}
\begin{figure}[t]
    \centering
    \includegraphics[width=0.99\linewidth]{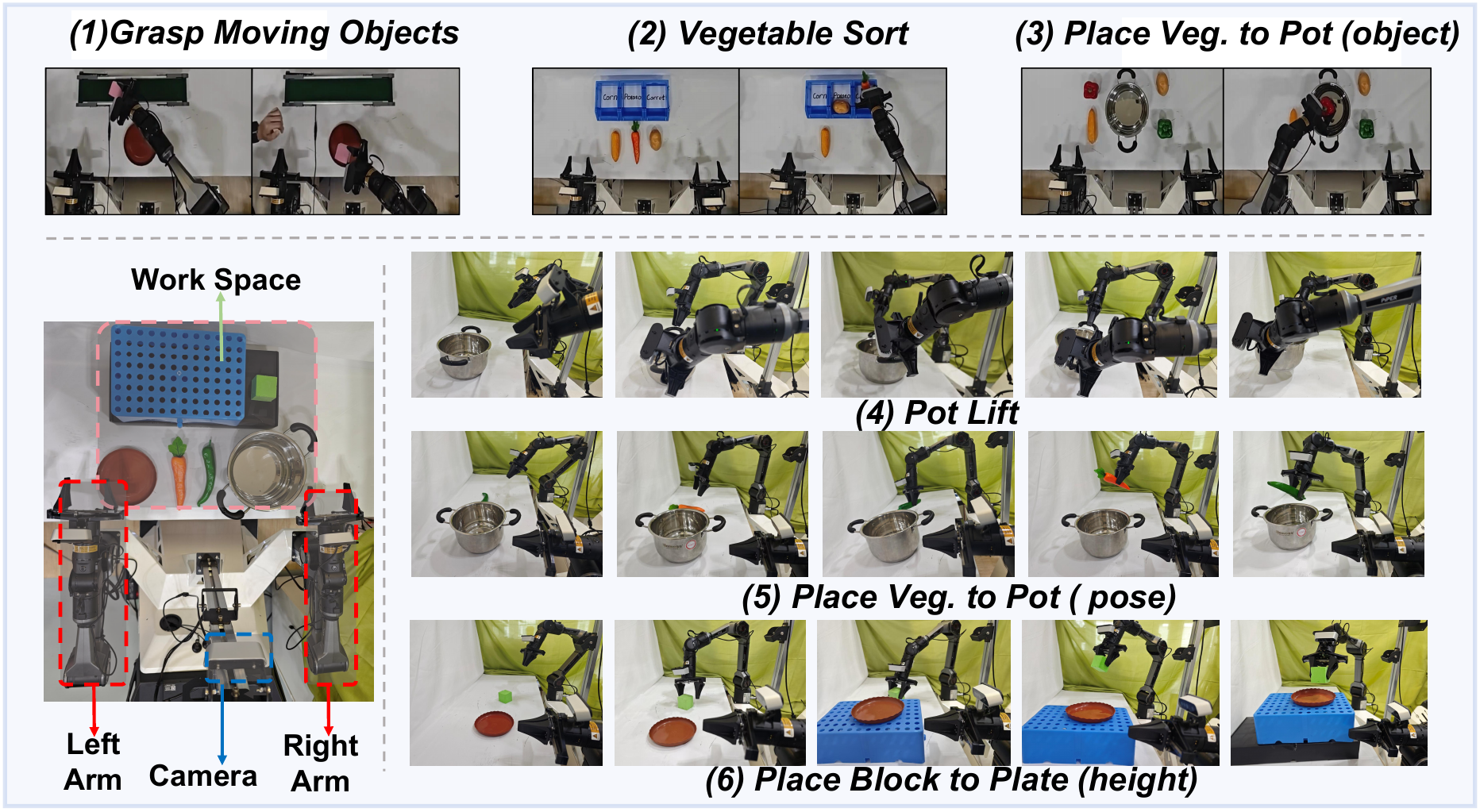}
    \caption{The real-world experimental setup and demonstrations of manipulation tasks.}
    \label{fig:real}
    \vspace{-0.2cm}
\end{figure}
\begin{table}[t]
    \centering
    \footnotesize
    \setlength{\tabcolsep}{4pt}
    \caption{Ablation on training data scale on CALVIN.}
    % \vspace{-0.3cm}
    \label{tab:ablation_data_scale}
    \begin{tabular}{c|ccc}
        \toprule
        Data Fraction & AR & DD & DFM (Ours) \\
        \midrule
        10\%  & 1.71 & 2.84 & \textbf{3.21}   \\
        50\%  & 3.01 & 3.88 & \textbf{4.03}   \\
        100\% & 4.18 & 4.32 & \textbf{4.58}   \\
        \bottomrule
    \end{tabular}
    \vspace{-0.3cm}
\end{table}

\Cref{tab:ablation_data_scale} shows that \method~consistently outperforms both autoregressive and discrete diffusion baselines across all data scales on CALVIN ABCD$\rightarrow$D.
At 10\% data, \method~achieves an Avg. Len. of 3.21, outperforming AR and DD methods by 1.50 and 0.37, respectively.
It also achieves the best result at both 50\% and 100\% data. 
% The particularly large margin in the low-data regime may be partly attributed to action noising during training: sampling multiple corrupted states from each action sequence exposes the model to diverse training inputs and thereby provides implicit data augmentation.
These results indicate that DFM-VLA is particularly beneficial in low-data regimes while maintaining consistent advantages as training data scales up.

\subsection{Real-World Experiments (RQ4)}
\label{sec:real}

\begin{table}[t]
    \centering
    \scriptsize
    \setlength{\tabcolsep}{2.5pt}
    \caption{Real-world success rates (\%) across
    six tasks. Baselines include $\pi_0$-FAST
    \citep{pertsch2025fast}, Dream-VLA
    \citep{yedreamVLA}, and $\pi_{0.5}$
    \citep{intelligence2025pi_}.}
    \label{tab:real_world_results}
    \resizebox{\linewidth}{!}{%
    \begin{tabular}{lccc
        >{\columncolor[gray]{0.9}}c
        >{\columncolor[gray]{0.9}}c}
        \toprule
        Task & $\pi_0$-FAST & Dream-VLA & $\pi_{0.5}$
        & \shortstack{\method\\w/o Embed}
        & \textbf{\method} \\
        \midrule
        \shortstack[l]{Grasp Moving Objects}
        & 50.0 & 57.5 & \textbf{82.5} & 70.0 & 75.0 \\
        \shortstack[l]{Sort Vegetables}
        & 47.5 & 55.0 & \textbf{70.0} & 65.0
        & \textbf{70.0} \\
        \shortstack[l]{Place Veg. to Pot (Object Var.)}
        & 60.0 & 67.5 & 72.5 & 77.5 & \textbf{82.5} \\
        Pot Lift
        & 50.0 & 57.5 & 65.0 & 70.0 & \textbf{77.5} \\
        \shortstack[l]{Place Veg. to Pot (Pose Var.)}
        & 42.5 & 62.5 & \textbf{72.5} & 67.5 & 70.0 \\
        \shortstack[l]{Place Block to Plate (Height Var.)}
        & 35.0 & 42.5 & 62.5 & 60.0 & \textbf{65.0} \\
        \midrule
        Average
        & 47.5 & 57.1 & 70.8 & 68.3 & \textbf{73.3} \\
        \bottomrule
    \end{tabular}%
    }
    \vspace{-0.3cm}
\end{table}
% \begin{table}[t]
%     % TODO: The first three task columns for \pi_0-FAST, Dream-VLA, and \method w Head
%     % are placeholder values and must be replaced with measured results.
%     \centering
%     \footnotesize
%     \setlength{\tabcolsep}{3pt}
%     \caption{Real-world success rate comparison (\%) on six tasks. The average is computed across all six tasks.}
%     \label{tab:real_world_results}
%     \resizebox{0.95\linewidth}{!}{%
%     \begin{tabular}{lcccccc|c}
%         \toprule
%         Method & \shortstack{Grasp Moving\\Objects} & \shortstack{Sort\\Vegetables} & \shortstack{Place Veg. to Pot\\(Object Variation)} & Pot Lift & \shortstack{Place Veg. to Pot\\(Pose Variation)} & \shortstack{Place Block to Plate\\(Height Variation)} & Average \\
%         \midrule
%         $\pi_0$-FAST~\citep{pertsch2025fast} & 50.0 & 47.5 & 60.0 & 50.0 & 42.5 & 35.0 & 47.5 \\
%         Dream-VLA~\citep{yedreamVLA} & 57.5 & 55.0 & 67.5 & 57.5 & 62.5 & 42.5 & 57.1 \\
%         $\pi_{0.5}$~\citep{intelligence2025pi_} & \textbf{82.5} & \textbf{70.0} & 72.5 & 65.0 & \textbf{72.5} & 62.5 & 70.8 \\
%          \rowcolor[gray]{0.9}
%         \textbf{\method~w/o embed} & 70.0 & 65.0 & 77.5 & 70.0 & 67.5 & 60.0 & 68.3 \\
%         \rowcolor[gray]{0.9}
%         \textbf{\method~} & 75.0 & \textbf{70.0} & \textbf{82.5} & \textbf{77.5} & 70.0 & \textbf{65.0} & \textbf{73.3} \\
%         \bottomrule
%     \end{tabular}%
%     }
%     \vspace{-0.3cm}
% \end{table}

\textbf{Setup.}
As shown in \Cref{fig:real}, 
we conduct real-world experiments on a bimanual AgileX platform equipped with two robotic arms, each with six degrees of freedom and a parallel gripper.
The system is instrumented with three RGB cameras: one fixed camera mounted at an elevated central viewpoint and two wrist-mounted cameras, one on each arm.

\noindent
\textbf{Task Setting.}
We design six representative manipulation tasks:
(1) grasping moving objects from a conveyor (\textit{Grasp Moving Objects});
(2) sorting different vegetables (\textit{Sort Vegetables});
(3) placing different vegetables into a pot (\textit{Place Veg. to Pot (Object Variation)});
(4) collaboratively lifting a pot with both arms (\textit{Pot Lift});
(5) placing vegetables with varying poses into a pot (\textit{Place Veg. to Pot (Pose Variation)});
and (6) placing a block onto a plate of varying height (\textit{Place Block to Plate (Height Variation)}).
For each task, we collect 100 training trajectories.
During evaluation, we conduct 40 trials per task and report the success rate.

\noindent
\textbf{Results.}
We compare \method~against representative methods
from three action-generation paradigms for
real-world manipulation: the autoregressive
baseline $\pi_0$-FAST, the discrete diffusion
baseline Dream-VLA, and the continuous
flow-matching baseline
$\pi_{0.5}$~\citep{intelligence2025pi_}.
As shown in~\Cref{tab:real_world_results},~\method~w/o
embed achieves an average success
rate of 68.3\%. In comparison, the
embedding-guided \method~improves the average by
5.0 percentage points to 73.3\%, outperforming the
strongest baseline, $\pi_{0.5}$, by 2.5 percentage
points. In particular,~\method~achieves the
highest success rates on \textit{Place Veg. to Pot
(Object Variation)}, \textit{Pot Lift}, and
\textit{Place Block to Plate (Height Variation)},
while remaining competitive with $\pi_{0.5}$ on
the other tasks. 
This advantage on tasks requiring semantic
understanding may stem from our discrete decoding
formulation. By sharing the VLM's discrete output
space, it can better leverage semantic priors
inherited from pretraining.
% These results indicate that the
% embedding-guided velocity-field construction
% provides stronger overall performance across
% diverse manipulation settings, with improvements
% distributed across multiple tasks rather than
% concentrated in a single task.

This result also demonstrates that traditional discrete
baselines (i.e., $\pi_0$-FAST and Dream-VLA) are
weaker than the continuous flow-matching baseline
$\pi_{0.5}$ because wrongly decoded action tokens
cannot be refined, leading to error accumulation.
In contrast, our method uses action velocities over discrete action tokens to
correct errors in time.
% Meanwhile, its unified discrete input-output space
% better preserves semantic understanding from
% autoregressive VLM backbones, yielding the best
% overall performance.
These results further show that action refinement
with discrete flow matching remains highly
competitive and robust in real-world settings.

\section{Conclusion}

We introduce \method, a discrete-flow-matching VLA that revises entire action sequences through token-level velocities. 
We study auxiliary-head and embedding-guided velocity-field constructions and introduce a Metric-Aligned Action Tokenizer for coarse-to-fine refinement. 
We employ a two-stage decoding strategy that combines iterative refinement with deterministic validation. 
Experiments across simulated and real-world tasks demonstrate strong performance, establishing discrete flow matching as a practical new decoding paradigm for VLA models.

% In this paper, we presented \method, a discrete flow matching framework for the VLA model that enables iterative full-sequence action refinement. Unlike autoregressive and discrete diffusion baselines, our \method~revisits and selectively updates action tokens through token-level velocity modeling, thereby improving action consistency in robotic manipulation.

% Our approach explores two velocity-field constructions: an embedding-guided formulation that defines structured probability paths and an auxiliary velocity head formulation that predicts velocities directly from hidden states. It further integrates kinetic-optimal velocity design with a practical two-stage decoder that couples iterative refinement and deterministic validation, achieving superior execution quality and favorable inference efficiency across CALVIN, LIBERO, and real-world evaluations.

% \textbf{Future work.} 
% Future work includes scaling DFM-based VLA pretraining, improving  token refinement, and extending the framework to tighter joint modeling of perception, reasoning, and action in more open-ended real-world environments.

\bibliographystyle{assets/plainnat}
\bibliography{AnonymousSubmission2027}

\end{document}